\documentclass[letterpaper, 10 pt, conference]{ieeeconf}

\IEEEoverridecommandlockouts 

\usepackage[utf8]{inputenc}
\usepackage{graphicx} 
\usepackage{caption}
\usepackage{subcaption}
\usepackage{xcolor}
\usepackage{hyperref}

\DeclareCaptionFormat{myformat}{\fontsize{8}{8}\selectfont#1#2#3}
\usepackage{cite}
\usepackage{amsmath,amssymb,amsfonts}
\usepackage{algorithmic}
\usepackage[ruled,vlined]{algorithm2e}
\SetKwInOut{Input}{input}
\SetKwInOut{Output}{output}
\SetKwInOut{Parameter}{parameters}

\usepackage{float}
\usepackage{textcomp}
\PassOptionsToPackage{hyphens}{url}
\usepackage{dblfloatfix}
\usepackage{listings,multicol}

\usepackage{multirow}

\title{\LARGE \bf
Towards Precise Pruning Points Detection using Semantic-Instance-Aware Plant Models for Grapevine Winter Pruning Automation
}

\author{Miguel Fernandes$^{1,2}$, Antonello Scaldaferri$^{1}$, Paolo Guadagna$^{3}$, Giuseppe Fiameni$^{4}$,\\ Tao Teng$^{1,3}$, Matteo Gatti$^{3}$ Stefano Poni$^{3}$, Claudio Semini$^{5}$,~\IEEEmembership{Member,~IEEE}, \\Darwin Caldwell$^{1}$,~\IEEEmembership{Senior Member,~IEEE}, Fei Chen$^{6}$,~\IEEEmembership{Senior Member,~IEEE}

\thanks{This research is supported by the project ``Grapevine Recognition, Manipulation and Winter Pruning Automation'' funded by IIT-Unicatt Joint Lab. The study was co-funded by Italian Ministry of University and Research PRIN 20172HHNK5 Project. \textit{(Corresponding author: Fei Chen)}}
\thanks{$^{1}$Miguel Fernandes, Antonello Scaldaferri, Tao Teng, Darwin Caldwell are with Active Perception and Robot Interactive Learning Laboratory, Department of Advanced Robotics, Istituto Italiano di Tecnologia, Via Morego 30, 16163, Genova, Italy (e-mail: {\tt\small name.surname@iit.it}).}%
\thanks{$^{2}$Miguel Fernandes is with Department of Informatics, Bioengineering, Robotics and System Engineering, Università di Genoa, Viale Causa 13, 16145 Genova, Italy (e-mail: {\tt\small miguel.ferreira@iit.it}).}%
\thanks{$^{3}$Paolo Guadagna, Tao Teng, Matteo Gatti, Stefano Poni are with Department of Sustainable Crop Production, Università Cattolica del Sacro Cuore, Via Emilia Parmense 84, 29122 Piacenza, Italy (e-mail: {\tt\small name.surname@unicatt.it}).}%
\thanks{$^{4}$Giuseppe Fiameni is with NVIDIA AI Technology Center (NVAITC), Italy (email: {\tt\small gfiameni@nvidia.com}).}
\thanks{$^{5}$Claudio Semini is with Dynamic Legged Systems (DLS) lab, Istituto Italiano di Tecnologia, Via Morego 30, 16163, Genova, Italy (e-mail: {\tt\small name.surname@iit.it}).}%
\thanks{$^{6}$Fei Chen is with Department of Mechanical and Automation Engineering, T-Stone Robotics Institute, The Chinese University of Hong Kong, Chung Chi Rd, Ma Liu Shui, Hong Kong (e-mail: {\tt\small f.chen@ieee.org}).}%
}

\begin{document}
\setlength{\textfloatsep}{0.1cm}
\setlength{\floatsep}{0.1cm}
\maketitle

\begin{abstract}
Grapevine winter pruning is a complex task, that requires skilled workers to execute it correctly.
The complexity makes it time consuming. It is an operation that requires about 80-120 hours per hectare annually, making an automated robotic system that helps in speeding up the process a crucial tool in large-size vineyards.
We will describe (a) a novel expert annotated dataset for grapevine segmentation, (b) a state of the art neural network implementation and (c) generation of pruning points following agronomic rules, leveraging the simplified structure of the plant.
With this approach, we are able to generate a set of pruning points on the canes, paving the way towards a correct automation of grapevine winter pruning.
\end{abstract}

\section{INTRODUCTION}
An important task to perform in a vineyard is winter pruning, a complex operation that needs to be completed during the dormant season \cite{PONI201688}. Performing a balanced winter pruning allows a good compromise between remunerative yield and desired grape quality hence maximizing grower's income \cite{Intrieri116, PONI2018445}. This selective operation requires annually about 80-120 man-hours per hectare. Due to increased skill shortage and limited labor availability in the agricultural sector, automating winter pruning is a crucial step to increase efficiency and reduce production costs.
\begin{figure}[t]
    \centering
    \includegraphics[width=0.43\textwidth]{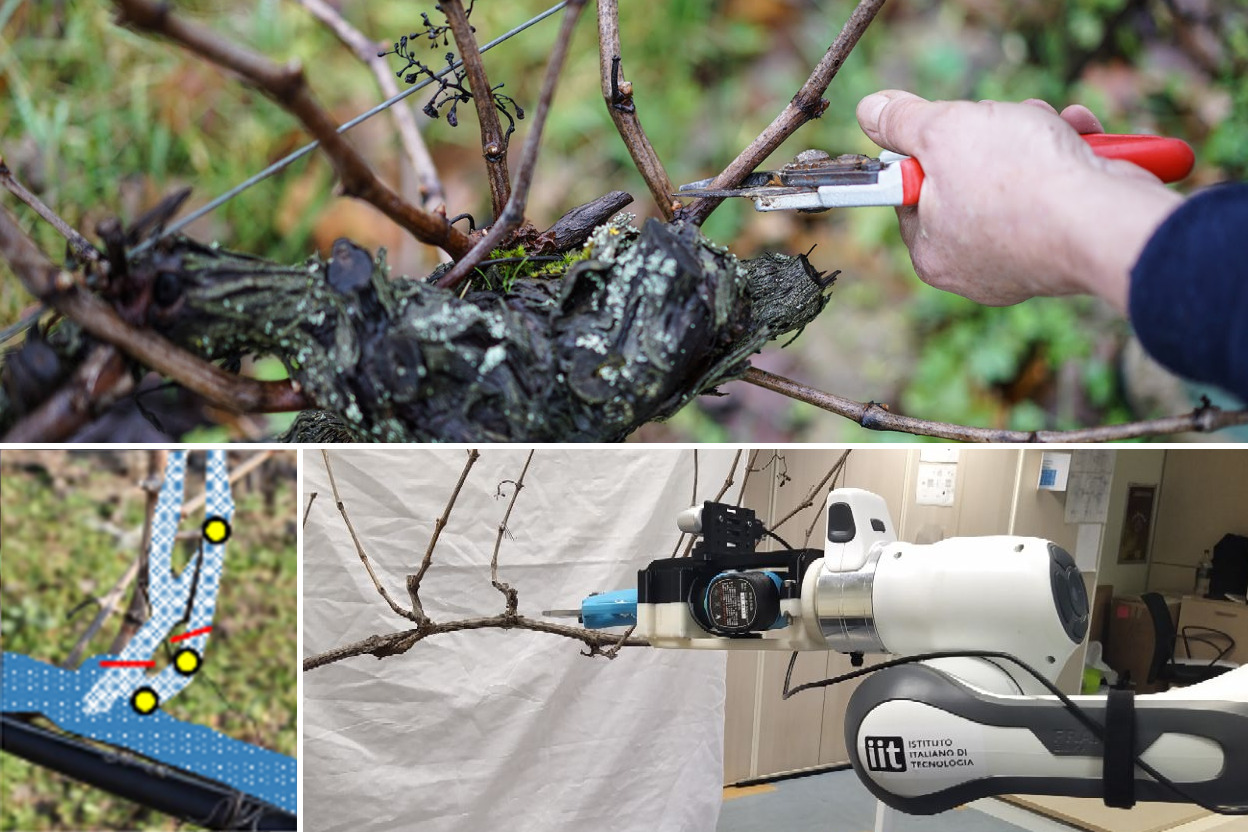}
    \caption{The top half shows an example of manual grapevine pruning and the bottom half illustrates our robot performing grapevine pruning along with the desired plant modeling, with the dark blue representing the main cordon, light blue the canes, yellow the nodes and red mark the pruning points.}
    \label{fig:intro:pruning}
\end{figure}

\begin{figure*}[t]
    \vspace{5pt}
    \centering
    \includegraphics[width=0.70\textwidth]{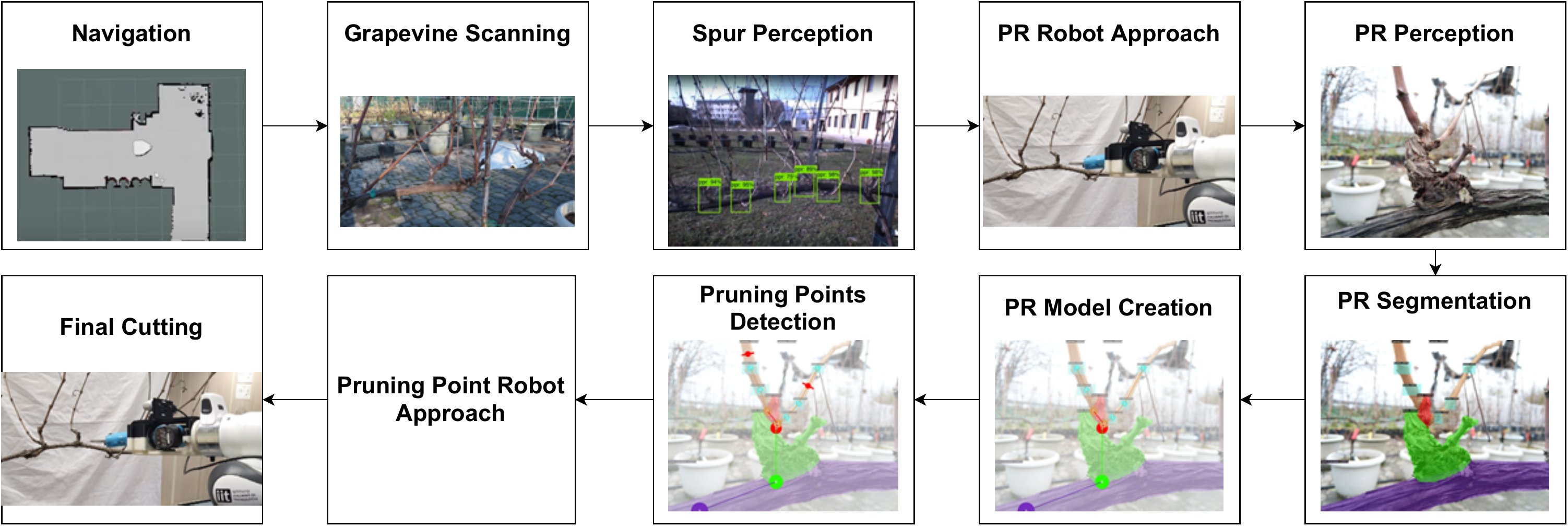}
    \caption{Pipeline of the winter pruning system, from the initial step of navigation up to the cutting.}
    \label{fig:intro:pipeline}
    \vspace{-18pt}
\end{figure*}
As part of the ongoing collaboration between Istituto Italiano di Tecnologia (IIT) and the Universita' Cattolica del Sacro Cuore, Piacenza (UCSC), the project VINUM has been addressing the automation of spur winter pruning of grapevines for the last few years. Figure \ref{fig:intro:pruning} shows a few images of hand pruning and a robot-performed pruning. Our current prototype has successfully performed several trials in controlled environments both inside the laboratory as well as in a simulated vineyard at UCSC (simulated means that the plants are growing in pots instead of the field)\cite{teng21icarm,guadagna21ecpa}. Figure \ref{fig:intro:pipeline} provides an overview of the pipeline that we developed, showing the sequence of operations that the pruning robot needs to execute from the approach to the vine to the extraction of the pruning points and cuttings. With consideration to our previous work\cite{prevpaper}, where we presented an initial dataset where grapevine organs were classified into 3 classes, a representative computer model of the grapevine specimen and an initial pruning point selection, we built upon those bases to create the new contributions of this paper:
\begin{itemize}
    \item A new dataset for grapevine segmentation based on five classes instead of three.
    \item A more uniform image capture process, along with its validation.
    \item A pruning points generation based on agronomic rules instead of proof of concept selection presented previously.
\end{itemize}
The neural network has also been retrained with the new dataset. In order to demonstrate the evolution, the neural network of our previous work is compared with the current one, along with the pruning points that are generated.

The remainder of this paper is structured as follows: Section \ref{sec:relwork} presents the related work, Section \ref{sec:dataset} describes the created dataset, Section \ref{sec:neuralnet} presents the description and training procedure for the neural network, Section \ref{sec:modeling} shows how we improved the grapevine plant model and the algorithm that creates the model. 
Section \ref{sec:ppgen} presents the improvement to the pruning point detection. The experimental setup and results are discussed in Section \ref{sec:results} and finally, the conclusions and possible future works are presented in Section \ref{sec:conclusions}.

\section{Related Work}
\label{sec:relwork}
In the deep learning field, various object segmentation\cite{Nogues2018}, detection\cite{Zhang2018}, and tracking algorithms\cite{Santos2019} have been presented in the scientific communities in the past years. Some of these studies have also been applied on the agri-food field, such as fruit detection \cite{Borianne2019} for yield estimation purposes \cite{Bargoti2016}, weed removal \cite{DiCicco2016, Milioto2017} and plant phenotyping \cite{Grimm2018}. The authors of \cite{doi:10.1002/rob.21680} developed a method that generates potential pruning points in long-cane pruned grapevines.
The authors in \cite{DiCicco2016} deal with the issues created by the variety of work regions in the fields, lighting, weather conditions, which leads to the difficulties for semantic segmentation of crop fields.
The authors in \cite{Grimm2018} present a proof of concept for quantifying and detecting plant organs for non-destructive yield estimation. This approach is based on automated detection, localization, count and analysis of plant parts used to estimate yield.
The authors in \cite{Santos2019} present a new public dataset with grape clusters annotated in 300 images and a new annotation with interactive image segmentation to generate object masks, identifying background and occluding foreground pixels using Scribbles. There is also an evaluation of two state-of-the-art methods for object detection, object segmentation and a fruit counting methodology. 
Authors in \cite{doi:10.1002/rob.21680} used a trinocular stereo camera system and correspondence algorithms to obtain a 3D reconstruction of the plant, which is useful to compute the pruning points in the wild. It achieves this by using a vehicle that encloses the grapevines, creating consistent lighting and background conditions that simplify the vision algorithms.

These representative works demonstrated several important concepts, such as real-time plant segmentation \cite{Milioto2017}, the more in-depth plant feature extraction for finding multiple parts of the plant\cite{Grimm2018}, and an annotation tool to generate a segmentation dataset quickly\cite{Santos2019} and a pruning generation system.
However, an approach to prune without the use of complex stereo camera systems or the use of mobile platform that encloses the grapevine, is still missing in the literature. Our proposed method uses an off-the-shelf depth camera (Intel realsense) and does not require a uniform background behind the grapevine.

\begin{figure}[tp]
    \centering
    \includegraphics[width=0.38\textwidth]{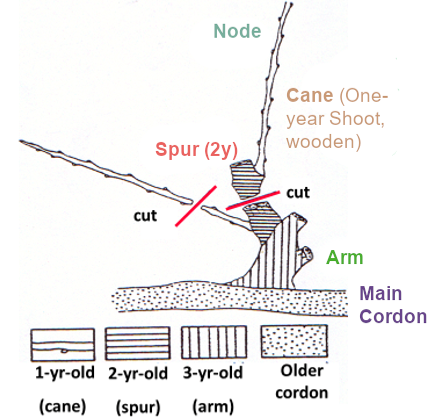}
    \caption{Schematic representation of a grapevine pruning region area, illustrating the 5 grapevine organs: Main Cordon, Arm, Spur, Cane, and Node. The red lines indicate the desired pruning points.}
    \label{fig:dataset:plantorgans}
\end{figure}

\section{Dataset}
\label{sec:dataset}
\begin{figure}[tpb]
    \centering
    \vspace{5pt}
    \includegraphics[width=0.40\textwidth]{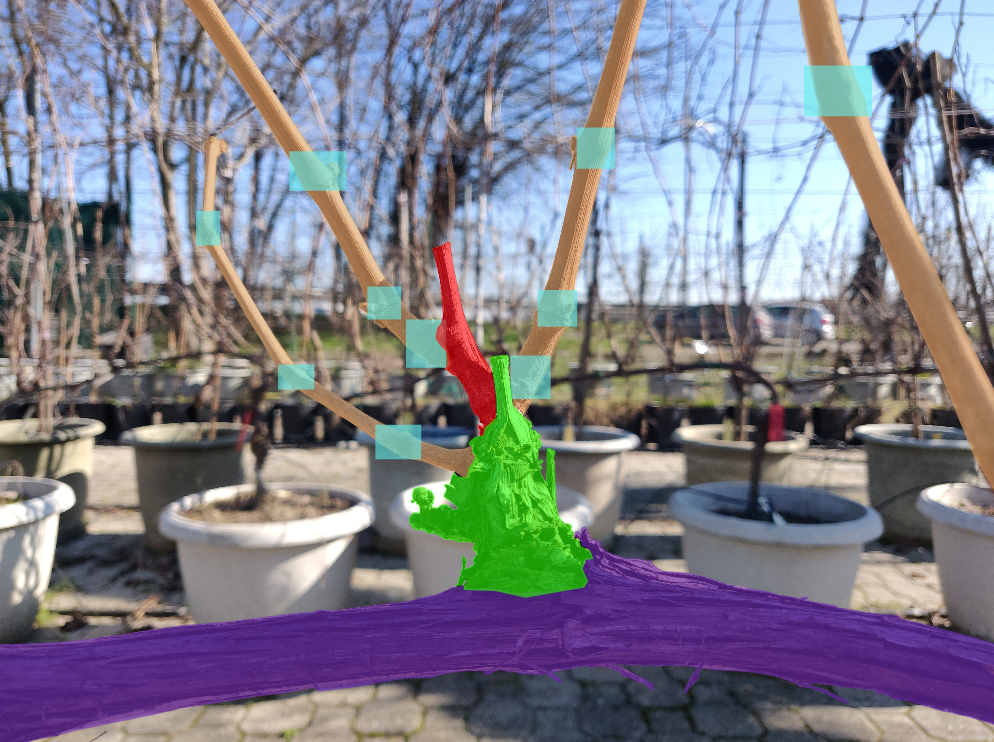}
    \caption{Example of a grapevine organ annotation, where purple represents the main cordon class, green the node arm class, red the spur class, orange the cane class, and light blue the node class. This picture is taken inside the simulated vineyard at UCSC, Piacenza, Italy.}
    \label{fig:dataset:annotation_exmpl}
\end{figure}

A dataset is an important component on the implementation of a machine learning method. Since no dataset with the plant organ segmentation of spur-pruned grapevines was found by the authors, it was decided to create a new dataset with the annotation of five classes: main cordon, arm, spur, cane and node.
This structure can be seen in the Fig. \ref{fig:dataset:plantorgans}, along with the correct pruning points for the shown example.
Each class corresponds to a different visible organ during the winter pruning.
The grapevine has several organs, the \textit{Main Cordon}, also referred as the permanent cordon.
Connected to the main cordon we may have at least three-year-old wood, which commonly would be called an \textit{Arm}.
Two-year-old wood which is the \textit{Spur}, the one-year-old structure is the \textit{Cane}.
The \textit{Cane} has several structures which are the \textit{Nodes}.
An example of these annotation concepts can be seen in Fig. \ref{fig:dataset:annotation_exmpl}.
Since  it is one of the main contributions to the scientific community, the authors shared the created dataset in open access format at \url{https://zenodo.org/record/5501784}.

\subsection{Data Acquisition}
\label{sec:datatset:aquisition}
The data acquisition was performed in the simulated vineyard at UCSC, Piacenza, Italy.
It is composed of two rows of grapevine specimens, running a shoot thinning experiment, where a group of seven plants is acting as control group, and the second group of eight grapevines is subjected to shoot thinning.
Each plant has around five spurs, and the photos are taken from both sides of the grapevine specimen, meaning that there is one picture where the plant grows from left to right, and the other side where the plant grows from right to left. The resolution of the images is 4608x3456 pixels, and we have a total of 148 images.
\begin{table}[t]
\vspace{5pt}
\centering
\caption{Number of annotations per class category in each plant group.}
\begin{tabular}{l|l|l|}
\cline{2-3}
                                  & Control & Shoot Thinning \\ \hline
\multicolumn{1}{|l|}{Main Cordon} & 79      & 84           \\ \hline
\multicolumn{1}{|l|}{Cane}        & 440     & 341          \\ \hline
\multicolumn{1}{|l|}{Node}        & 110     & 912          \\ \hline
\multicolumn{1}{|l|}{Arm}         & 103     & 154          \\ \hline
\multicolumn{1}{|l|}{Spur}        & 116     & 144          \\ \hline
\multicolumn{1}{|l|}{Total}       & 1838    & 1635          \\ \hline
\end{tabular}
\end{table}
There are other two sets of images that were created, one that contains 100 images, captured from another row present in the simulated vineyard, with the same resolution as previously mentioned but currently not annotated. The other is an assorted group of images, containing total of 196 images, from various angles and various resolutions that is partly annotated, with 153 images annotated out of 196.

\subsection{Data Annotation}
\label{sec:dataset:annotation}
The dataset is being annotated following the COCO segmentation dataset, with the five mentioned classes, the main cordon, the arm, the spur, the cane, and the node.
We chose the COCO format for the annotation since it is a common format for segmentation annotation, which contributes to a higher availability of annotation tools that use this format, as well as some neural network frameworks provide built-in processing of this format.
The annotation tool being used is COCO annotator, a web-based tool that is designed for efficiently label images.
The 149 captured images were annotated and then split into 80\% for training and 20\% for images, corresponding to 119 training images and 30 evaluation images. We consider that there is no current need for a testing set since what can be considered the testing set for the neural network is the on-field testing of the platform.
The images are being annotated in their original size, allowing to downscale the image as required.
These images were annotated by two agronomy experts.

\section{Neural Network}
\label{sec:neuralnet}
This section describes the neural network architecture and backbone that was used for this work.
The framework that was used is the same as the previous work, which is Detectron2\cite{wu2019detectron2}, 
Facebook AI Research's implementation of state-of-the-art detection algorithms, using Pytorch, using Mask R-CNN, and, instead of using several backbones, we settled for just one, which is the ResNext X101-FPN\cite{DBLP:journals/corr/XieGDTH16}. 
This ResNext is an improvement on the original ResNet network, considering a new dimension named cardinality on top for the normal height and depth of a neural network, which is the size of the set of transformations.

The network is being trained using the default training procedure of Detectron2. This procedure creates a model, optimizer, scheduler, and dataloader with the default configurations provided along with the model. It then loads the pre-trained model weights, initializes logging functions and starts to follow a standard training workflow with a single-optimizer single-datasource iterative optimization. The training hyperparameters are the default ones, with the only changes being the batch size changed to 2, from original value 16, and the number of training iterations that was changed to 50000, from the original 270000. These changes were made to adjust to the size of our dataset, that is considered small for the network itself.

\section{Grapevine Plant Model Structure \& Modeling Algorithm}
\label{sec:modeling}
In our previous work \cite{prevpaper}, we introduced a 2D plant modeling algorithm, which considers only three types of organs.
This type of model can lead to inconsistencies in the overall model, resulting in an unreliable information.
Understanding that there where ways to improve the model, using depth information, we present a new modeling algorithm, that we consider more reliable and that has stricter relationships between the different detected organs of the plant, considering the new five-class categorization.

\subsection{Grapevine Plant Model Structure}
To create this model we use a tree-based data structure, since its structure is suitable to represent the relationship of the different detected grapevine organs.
Each node of this data structure is a generic grapevine item, which is the computational representation of a grapevine organ. Each of them contains all the information required for the subsequent pruning points estimation step, such as \textit{organ type, instance ID, bounding box, segmentation mask, origin, endpoint, parent organ, children organs, distance from the parent organ} and other information used for graphical representation.
The relationship information is stored in the \textit{parent} and \textit{children} attributes of each node. In addition, the children items of each node are sorted by the value of the \textit{distance from the parent} parameter.
Based on the new five-class categorization, we defined a new set of more specific and stricter relationships between the different types of organs, to create the new plant model, listed in Table \ref{tab:connections_types}.
\begin{table}[t]
\vspace{5pt}
    \centering
    \caption{Relationships between the different types of organs.}
    \captionsetup{justification=centering}
    \begin{tabular}{|c|c|}
        \hline
        \textbf{Parent Organ Type} & \textbf{Children Organs Types}\\
        \hline
        \textit{Main Cordon} & \textit{Arm}, \textit{Spur}, \textit{Cane}\\
        \hline
        \textit{Arm} & \textit{Spur}, \textit{Cane}\\
        \hline
        \textit{Spur} & \textit{Cane}\\
        \hline
        \textit{Cane} & \textit{Node}\\
        \hline
    \end{tabular}
    \label{tab:connections_types}
\end{table}

\subsection{Grapevine Plant Modeling Algorithm}
For the modeling algorithm, we use an upgraded version of the previous algorithm in \cite{prevpaper} to compute the connections between two sets of organs of different types, whose pseudo-code is given in Algorithm \ref{alg:get_connections_algorithm}.
It uses the same concept of overlap between masks as the previous one and uses the same masks dilation and subdivision heuristics.
The main difference is that now we consider both upwards and downwards growing directions of the organs, i.e. organs that grow either from the upper side of the parent organ or from the lower side.
To compute all the final connections for the entire model, this algorithm is executed for each type of \textit{parent-children} connection, according to Table \ref{tab:connections_types}.
Each time, it takes as input the two corresponding sets of items, the first for items of type \textit{parent} and the second for items of type \textit{children}.
For the last type of connections, where the \textit{nodes} are considered, the set of the children masks is replaced with their corresponding bounding boxes for simplicity.
\begin{algorithm}[t]
\vspace{5pt}
    \SetAlgoVlined
    \Parameter{$dilation, max\_iter, n, include\_top$}
    \Input{$items_A \rightarrow{\text{Set of type A items}}$}
    \Input{$items_B \rightarrow{\text{Set of type B items}}$}
    \Output{$connections \rightarrow{\text{Map of connections}}$}
    \caption{Compute connections between two types of organs.}
    \captionsetup{justification=centering}
    $masks_A \gets$ set of masks of $items_A$\\
    $IDs_A \gets$ set of instance IDs of $items_A$\\
    \ForEach{$i, mask_A \in \textbf{enumerate}(masks_A)$}{
        $masks_A[i] \gets mask_A * (IDs_A[i] + 1)$}
    \ForEach{$item \in items_B$} {
        $iter \gets 0, connected \gets False$\\
        $mask \gets item$ segmentation mask\\
        \While{$iter \leq max\_it \textbf{ and not}(connected)$} {
            \If{$iter > 0$} {
                Dilate $mask$
            }
            $indices \gets mask$ non-zero values indices\\
            Obtain $masks_A$ unique values at $indices$\\
            \If{There are non-zero correspondences} {
                $ID_A, mask_A \gets [lowest]$ correspondence\\
                $inter \gets \textbf{intersection}(mask_A, mask)$\\
                \If{$inter$ is in the $mask$ $[n^{th}]$ slot} {
                    $item_A \gets items_A(ID_A)$\\
                    Add $(item_A, item)$ to $connections$\\
                    $connected \gets True$\\
                }
            }
            $iter++$
        }
        \If{$\textbf{not}(connected) \textbf{ and } include\_top$} {
                Repeat the previous \textbf{while} loop considering the $[highest]$ correspondence and the $[1^{st}]$ slot\\
        }
    }
    \label{alg:get_connections_algorithm}
\end{algorithm}

\section{Pruning Points Detection}
\label{sec:ppgen}
After obtaining the grapevine plant model, it is used to compute the pruning points.
One of the main contributions of this work is to provide a set of pruning points that is as reliable as possible, meaning, the points that are provided will not damage the plant if cut. There is the possibility that certain cuts will not be performed. 
In order to obtain such a set of points, we defined a set of assessments to simplify the reasoning process of the pruning points detection phase. The set is comprised of the following assessments: pruning region location, pruning region canes number, basal cane growth direction, basal cane vigor, basal cane origin and adjacent pruning regions distance.
In the created model, each pruning region is a grapevine item and its type can be either \textit{arm}, \textit{spur} or \textit{cane}, according to Table \ref{tab:connections_types}.
The pruning region location is the side of the \textit{main cordon} from which the pruning region grows up.
For its assessment, three different angles $\alpha_V$, $\alpha_I$ and $\alpha_D$ are considered to classify the pruning region as ventral, intermediate or dorsal, which means that the pruning region grows from the bottom, middle or top side, respectively.
We define $(x_{PR}, y_{PR})$ as the pruning region origin image coordinates and $d_{MC}$ as the diameter of the \textit{main cordon}, that can be estimated from its segmentation mask, at the column $x_{PR}$, in pixels.
Moreover we define $y$ as the first non-zero row in the \textit{main cordon} segmentation mask.
With these definitions, the pruning region location is classified as:
\begin{equation}
    \begin{cases}
        dorsal & \mbox{if } y_{PR} < y + \frac{d_{MC}}{2} * (1 - \cos{\frac{\alpha_D}{2}})\\
        ventral & \mbox{if } y_{PR} > y + d - \frac{d_{MC}}{2} * (1 - \cos{\frac{\alpha_V}{2}})\\
        intermediate & \mbox{otherwise}
    \end{cases}
    \label{eq:pr_type}
\end{equation}
For the pruning region canes number assessment, a Depth-first search algorithm is used, counting the items of type \textit{cane} in the sub-tree represented by the pruning region.
The basal cane, in a pruning region, is the \textit{cane} closest to the \textit{main cordon}.
For its growth direction assessment, other two different angles $\alpha_L$ and $\alpha_C$ are considered to establish whether it is vertical or not, in both the lateral and cross sections respectively.
We define $(x_{o}, y_{o}, z_{o})$ and $(x_{e}, y_{e}, z_{e})$ as the basal cane origin and endpoint in the 3D space, that can be computed thanks to the depth image and camera intrinsic parameters.
With these definitions, the growth direction can be classified as:
\begin{equation}
    \begin{cases}
        vertical & \mbox{if } \frac{|x_o - x_e|}{|y_o - y_e|} \leq \alpha_L \And \frac{|z_o - z_e|}{|y_o - y_e|} \leq \alpha_C\\
        not\ vertical & \mbox{otherwise}
    \end{cases}
\end{equation}
The vigor of the basal cane is its thickness and it can be estimated from its segmentation mask and the depth image.
It is used to establish which canes to keep and which to remove, if too thick or too thin, based on two threshold values.
We define $rows$ and $cols$ as the row and column indices of the basal cane segmentation mask.
We suppose that the basal cane is vertical or horizontal based on its bounding box aspect ratio, for simplicity.
The subsequent considerations are valid for vertical basal canes.
In the case of horizontal basal canes just swap the rows with the columns.
Since the depth at the basal cane segmentation mask boundaries may not be accurate, for each $row$ in $rows$, we consider an estimation of the $row$ depth, considering the midpoint between the columns of the mask in that specific $row$.
This $row$ depth estimation is then used to obtain the 3D points at the $row$ boundaries, to compute the $row$ thickness as:
\begin{equation}
    \begin{gathered}
        thicknesses \gets \mbox{empty list}\\
        \forall row \in rows\\
        cols_{row} \gets \mbox{get $cols$ values where $rows$ is $row$}\\
        col_{mean} \gets mean(cols_{row})\\
        depth \gets estimate\_real\_depth(\langle col_{mean}, row \rangle)\\
        col_m, col_M \gets \min(cols_{row}), \max(cols_{row})\\
        \vec{p}_{row_1} \gets to\_3D(\langle col_m, row \rangle, depth)\\
        \vec{p}_{row_2} \gets to\_3D(\langle col_M, row \rangle, depth)\\
        thickness_{row} \gets |\vec{p}_{row_1} - \vec{p}_{row_2}|\\
        \mbox{add $thickness_{row}$ to $thicknesses$}
    \end{gathered}
\end{equation}
The vigor final estimation is computed considering the mean value of all the $thicknesses$ among all the $rows$.

Another important point is the basal canes origin assessment, that is necessary for two evaluations.
The first is to classify the pruning region as new or not, verifying whether the \textit{parent organ type} is \textit{main cordon} or not, respectively.
The second is to verify if the pruning region concerns a replacement cut or not, verifying whether the parent organ type is \textit{arm} or not, respectively.
The last considered assessment is the adjacent pruning regions distance assessment.
This is necessary in the cases where there are multiple new pruning regions that are close together.
In this cases, based on a threshold value, it is verified whether these pruning regions are too close or not, in order to remove or keep them, respectively.
We define $PRs$ as the set of all the grapevine plant pruning regions. For each pruning region $PR$ in $PRs$, its distance $d_{PR}$ to adjacent pruning regions is computed as:

\begin{equation*}
    \begin{gathered}
        PR_{-1} \gets \mbox{get the previous pruning region from } PRs\\
        PR_{+1} \gets \mbox{get the next pruning region from } PRs\\
        \vec{p}_{PR} \gets \mbox{get the 3D origin of } PR\\
        \vec{p}_{PR_{-1}} \gets \mbox{get the 3D origin of } PR_{-1}\\
        \vec{p}_{PR_{+1}} \gets \mbox{get the 3D origin of } PR_{+1}\\
        d_{PR} \gets \max(|\vec{p}_{PR} - \vec{p}_{PR_{-1}}|, |\vec{p}_{PR} - \vec{p}_{PR_{+1}}|)
    \end{gathered}
\end{equation*}

Furthermore, in order to perform the final pruning in a correct manner, we defined different types of cuts for the grapevine organs, the details of which are given in Table \ref{tab:cuts_types}.
\begin{table}[t]
\vspace{5pt}
    \centering
    \caption{Types of pruning cuts.}
    \captionsetup{justification=centering}
    \begin{tabular}{|p{0.35\linewidth}|p{0.55\linewidth}|}
        \hline
        \multirow{1}{*}{\textbf{Cut type}}\centering & \textbf{Description}\\
        \hline
        \multirow{2}{*}{\textit{Clean cut}}\centering & The corresponding organ is completely removed.\\
        \hline
        \multirow{2}{*}{\textit{Base-bud cut}}\centering & The base-buds are kept only, while the remaining (upper) part is removed.\\
        \hline
        \multirow{4}{*}{\textit{Spur cut}}\centering & The first \textit{N} nodes of the basal cane are left, while the rest of it is removed. If it is not a new pruning region, the \textit{spur parent} organ is cut above the basal cane.\\
        \hline
        \multirow{3}{*}{\textit{Replacement cut}}\centering & The first \textit{N} nodes of the basal cane are left, while the rest of it is removed. The \textit{arm parent} organ is cut above the basal cane.\\
        \hline
    \end{tabular}
    \label{tab:cuts_types}
\end{table}
According to the assessments, evaluations, rules, heuristics, equations and types of cuts described above, the final set of pruning points is derived.
In particular, each pruning point $\vec{pp}$ is computed as a point between two other points $\vec{p}_1$ and $\vec{p}_2$.
These points $\vec{p}_1$ and $\vec{p}_2$ can be \textit{nodes} locations, \textit{canes} / \textit{spurs} / \textit{arms} origins or endpoints.
It may happen that the estimated pruning points are too far from their corresponding organ origin.
This can happen when \textit{nodes} are not detected.
In these cases, it is helpful to compute the pruning point using a distance parameter $d$, that can be set by the user, based on the vineyard grapevine type.
With these definitions, a pruning point $\vec{pp}$ can be computed as:
\noindent
\begin{equation*}
    \begin{gathered}
        D \gets |\vec{p}_1 - \vec{p}_2|\\
        w_1 \gets \frac{|D - d|}{D}\\
        w_2 \gets 1 - w_1\\
        \vec{pp} \gets w_1 * \vec{p_1} + w_2 * \vec{p_2}\\
    \end{gathered}
    \label{eq:pruning_point}
\end{equation*}
And its corresponding orientation angle $\alpha$ as:
\noindent
\begin{equation*}
    \begin{gathered}
        \delta_x \gets {p_1}_x - {p_2}_x\\
        \delta_y \gets {p_1}_y - {p_2}_y\\
        \alpha \gets \begin{cases}
        0, & \mbox{if } \delta_x = 0 \\
        \frac{\pi}{2}, & \mbox{if } \delta_y = 0 \\
        \arctan{\frac{\delta_y}{\delta_x}} - \mathop{\mathrm{sign}}{\frac{\delta_y}{\delta_x}} * \frac{\pi}{2}, & \mbox{otherwise}
        \end{cases}
    \end{gathered}
\end{equation*}
Note that if $d$ is $0$, the pruning point is placed in $\vec{p}_1$ and if $d$ is $\frac{1}{2}$, the pruning point is the midpoint between $\vec{p}_1$ and $\vec{p}_2$.

Since the pruning point is computed as a point between two other points, along a straight line, it may occur that it is not on the corresponding grapevine organ, that means it is not contained in the corresponding grapevine item segmentation mask. As such, we need to move the estimated points on their corresponding organs.
This correction is applied to each pruning point, moving the point horizontally or vertically until it reaches either its corresponding segmentation mask or a meaningful area of the depth image.
\begin{figure*}[t]
\vspace{5pt}
    \centering
    \begin{subfigure}{.26\textwidth}
        \centering
        \includegraphics[width=0.9\linewidth]{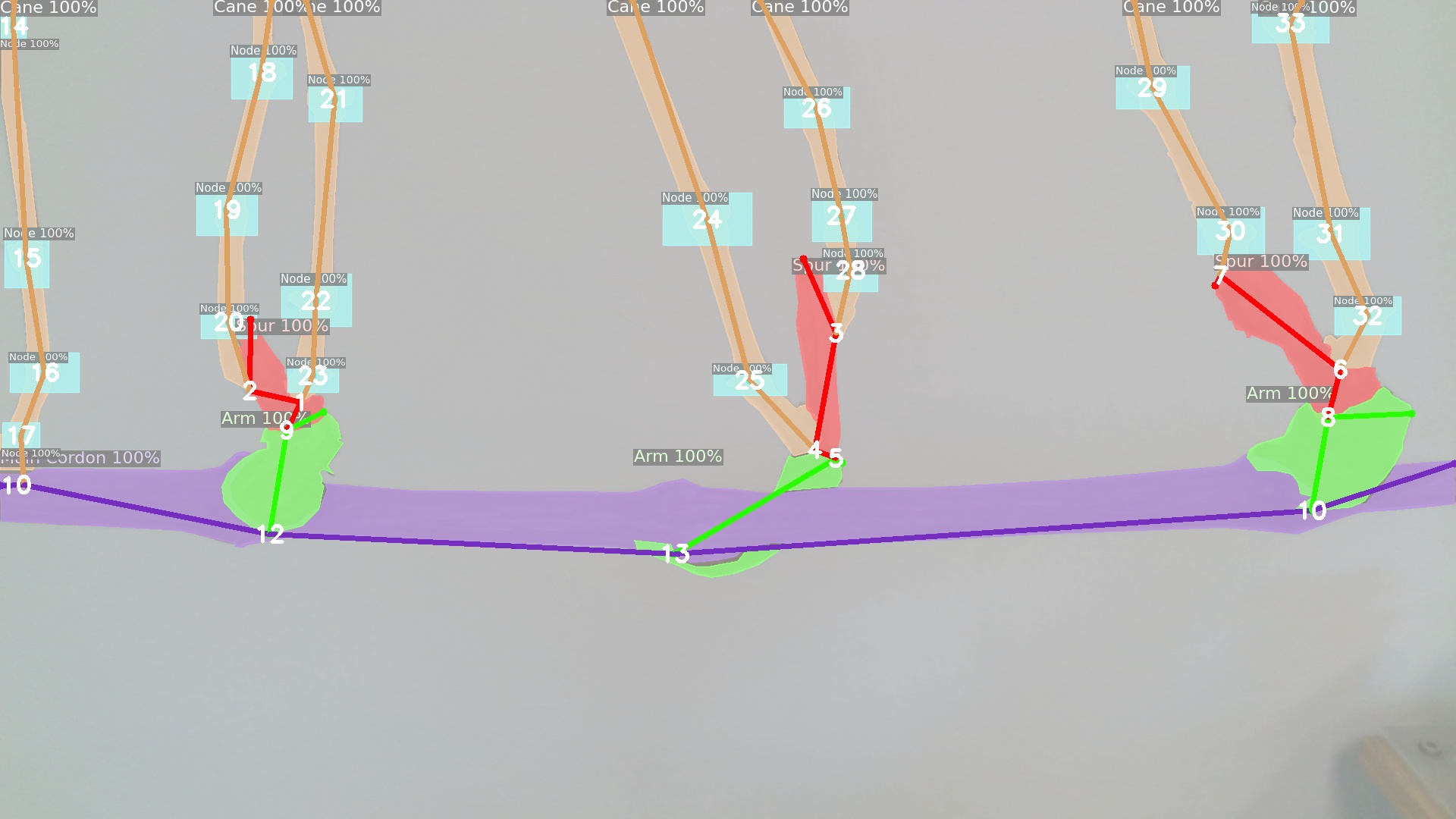}
        \caption{Ground-Truth Segmentation.}
        \label{fig:modeling_1}
    \end{subfigure}%
    \begin{subfigure}{.26\textwidth}
        \centering
        \includegraphics[width=0.9\linewidth]{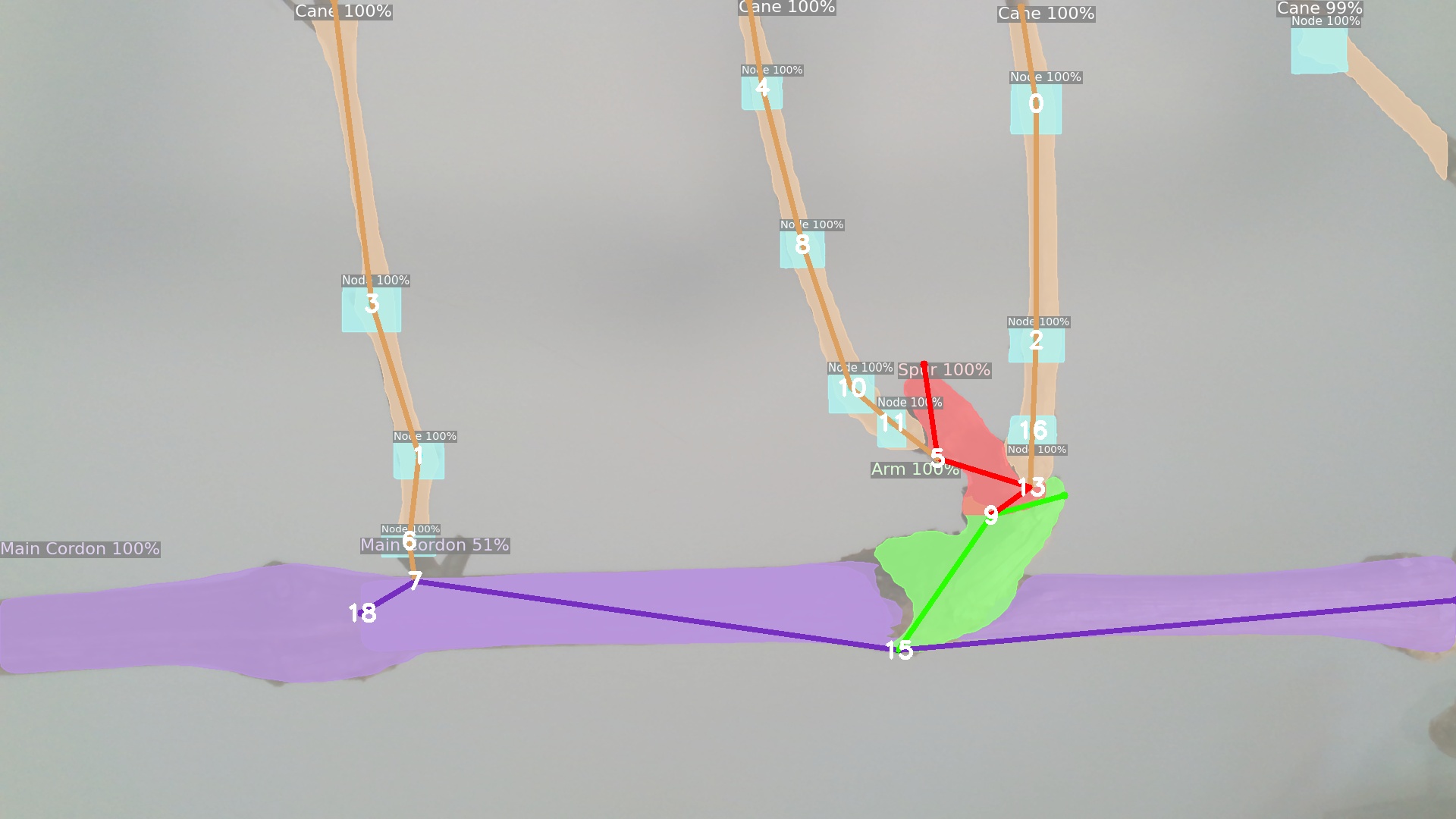}
        \caption{Inferred Segmentation in the lab.}
        \label{fig:modeling_2}
    \end{subfigure}%
    \begin{subfigure}{.26\textwidth}
        \centering
        \includegraphics[width=0.9\linewidth]{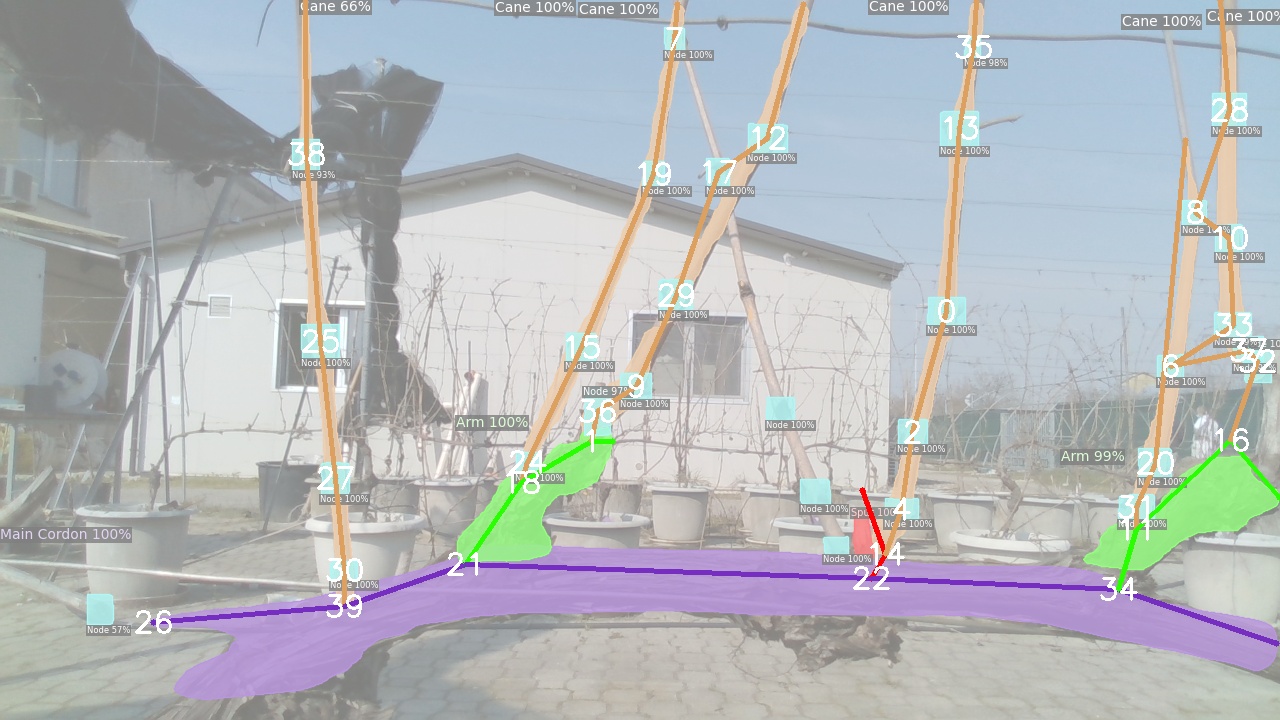}
        \caption{Inferred Segmentation in the wild.}
        \label{fig:modeling_3}
    \end{subfigure}
    \caption{Modeling Algorithm Results: in this figure, the models are drawn on the color image, to show that it contains all the information needed to compute the pruning points, using a simplified structure and discarding the complications of the raw images. Fig.\ref{fig:modeling_1} uses the ground-truth, while Fig. \ref{fig:modeling_2} uses the output of the neural network, with both cases being in the lab environment. Fig. \ref{fig:modeling_3} shows the segmentation neural network operating on the simulated vineyard.}
    \captionsetup{justification=centering}
    \label{fig:modeling}
    \vspace{-10pt}
\end{figure*}
\begin{figure*}[t]
    \centering
    \begin{subfigure}{.26\textwidth}
        \centering
        \includegraphics[width=0.9\linewidth]{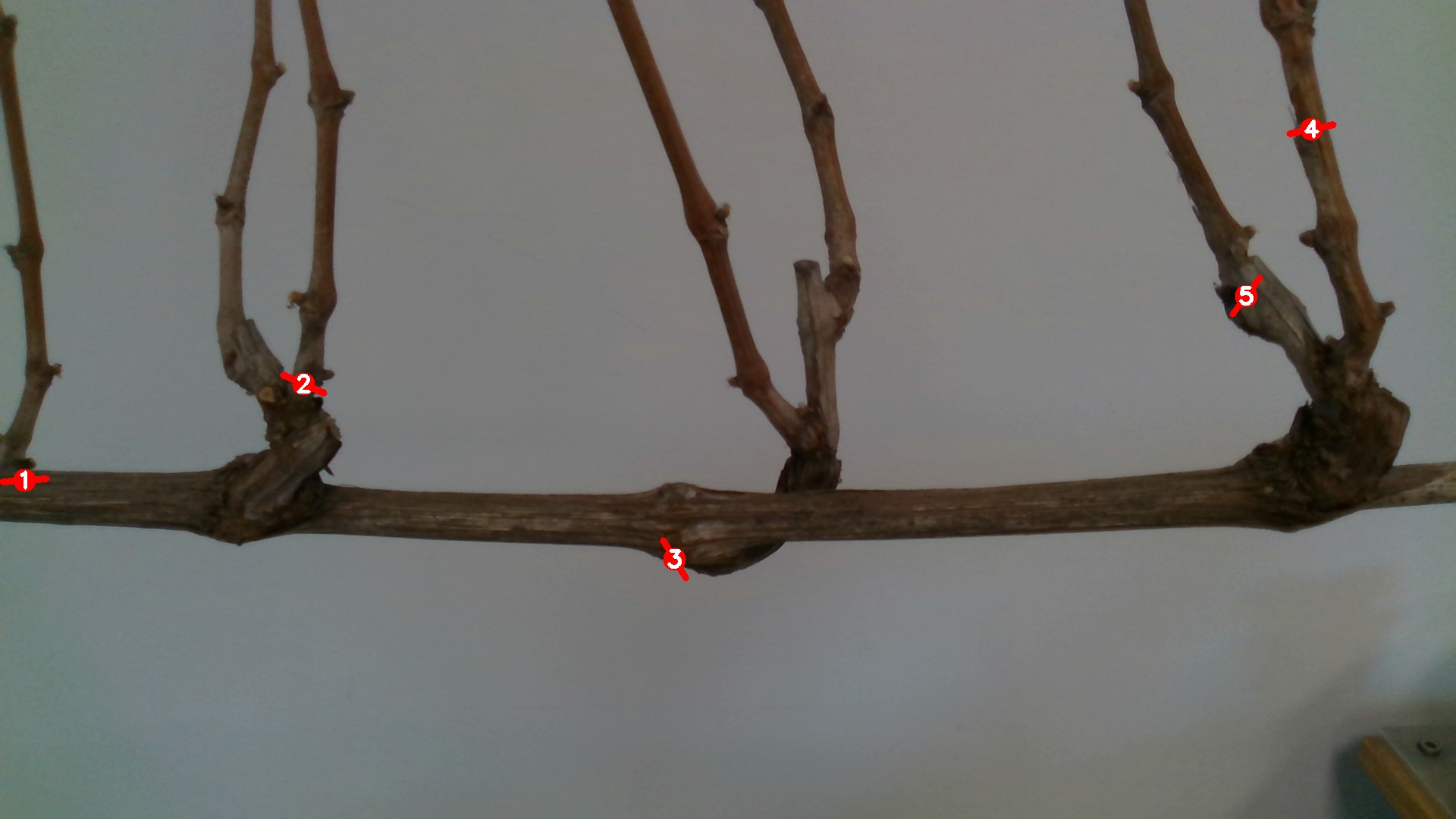}
        \caption{Ground-Truth Segmentation.}
        \label{fig:detection_1}
    \end{subfigure}%
    \begin{subfigure}{.26\textwidth}
        \centering
        \includegraphics[width=0.9\linewidth]{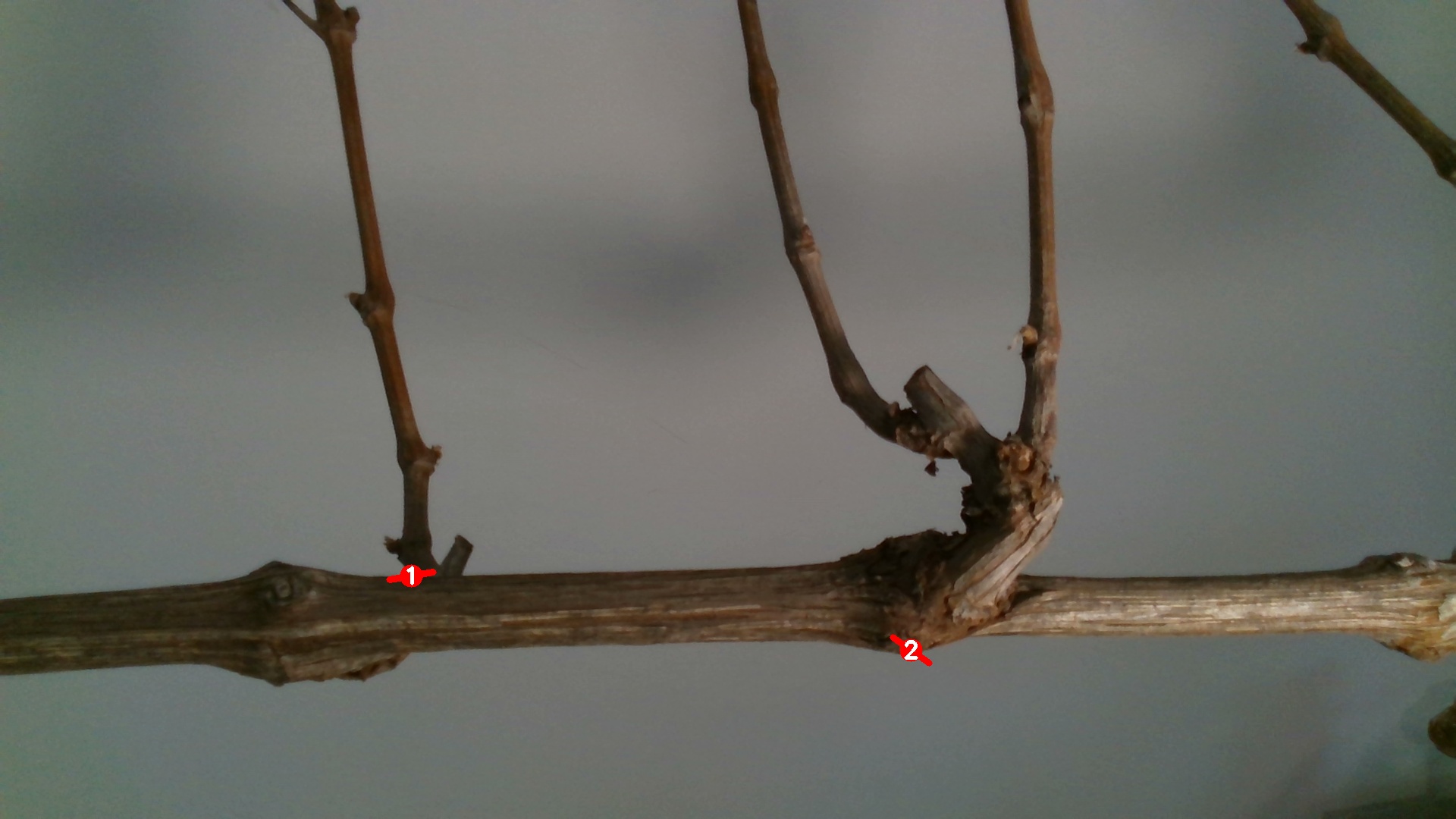}
        \caption{Inferred Segmentation in the lab.}
        \label{fig:detection_2}
    \end{subfigure}%
    \begin{subfigure}{.26\textwidth}
        \centering
        \includegraphics[width=0.9\linewidth]{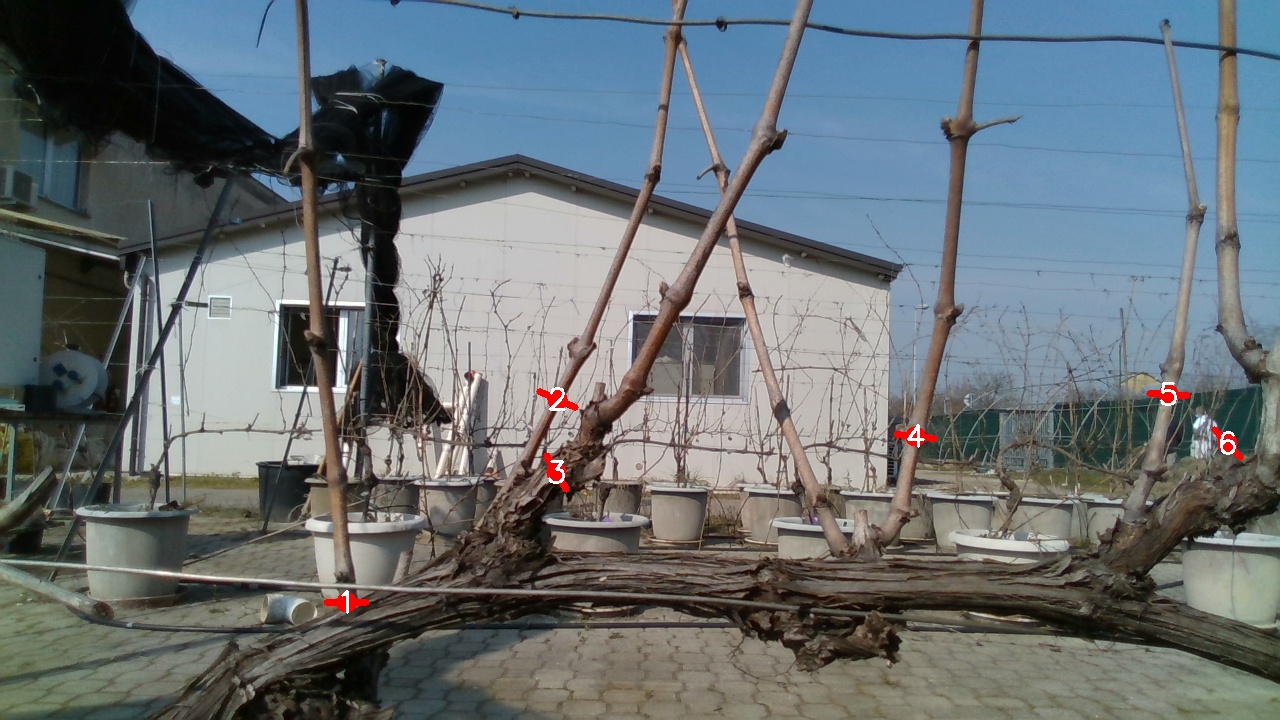}
        \caption{Inferred Segmentation in the wild.}
        \label{fig:detection_3}
    \end{subfigure}
    \caption{Pruning Points Estimation Results: in this figure, the corresponding pruning points are estimated using the plant models in Fig. \ref{fig:modeling}. In these images we can see how the pruning points locations accuracy becomes lower as we move from perfectly known cases, as in Fig. \ref{fig:detection_1}, to unknown and challenging cases, as in Fig. \ref{fig:detection_3}.}
    \captionsetup{justification=centering}
    \label{fig:detection}
    \vspace{-15pt}
\end{figure*}
\section{Experimental Setup \& Results}
\label{sec:results}
\subsection{Experimental Setup}
For the performance evaluation of the modeling and the pruning points detection algorithms, we considered two different setups.
The first setup, in a controlled environment, using either ground-truth 
or inferred
segmentations by the grapevine segmentation neural network, of grapevine plants pictures acquired inside the lab.
The second setup, is based on the simulated vineyard, using the segmentation masks and the detected grapevine items inferred by the grapevine segmentation neural network.

\subsection{Experimental Results}
Concerning the neural network, the evaluation metrics used are the intersection over union, as seen in Table \ref{tab:results:nntv}. We can see that there is a significant improvement over our previous results, but we still consider that there is room for improvement. With these results we also consider that the dataset can and is useful for grapevine organ segmentation, although for neural network training the dataset would improve by having more and varied samples.

\begin{table}[t]
    \caption{Average Precision and Average Recall after training comparing the previous work X101 backbone and the current X101 backbone. Both metrics are calculated with intersection over union IoU=0.50:0.90, considering all area sizes and a maximum detections value of 100. The dataset for this evaluation is composed of 60 images  obtained from the current dataset and the previous work dataset, updated to have the current five classes. }
    \label{tab:results:nntv}
    \begin{center}
        \begin{tabular}{|c|c|c|}
            \hline
            & Previous Work & Current Work\\
            \hline
            Average Precision & 27.5\% & 46.2\% \\
            \hline
            Average Recall & 30.8\% & 54.7\% \\
            \hline
        \end{tabular}
    \end{center}
\end{table}

The performance of both the modeling algorithm and the pruning points detection algorithm cannot be evaluated in a numerical manner, using conventional performance indices, because no ground-truths were created for either plant models or pruning points.
For this reason, this evaluation is based only on the qualitative aspects of the obtained results.

\subsubsection{Modeling Algorithm}

Figure \ref{fig:modeling} shows some output examples of the modeling algorithm.
We can notice how its performance strictly depends on the detection, recognition and segmentation inference of the grapevine segmentation neural network.
In Fig. \ref{fig:modeling_1}, where the segmentation masks are hand-labeled, we can see how the modeling algorithm is perfectly capable to reconstruct the plant model.
As the segmentation masks deteriorate, the plant modeling accuracy also decreases and this can be well noted in Fig. \ref{fig:modeling_3}, where we have a scene in-the-wild.
The modeling algorithm is not always perfect even with hand-labeled segmentations, because, if some organs are occluded, while they are well visible from the other side of the plant, it is not possible to establish the geometries of such pruning regions from a single point of view.
This shows that the plant modeling performance is highly dependent on the pruning regions points of view.

\subsubsection{Pruning Points Estimation}

Figure \ref{fig:detection} shows the pruning points estimated from the extracted models in Fig. \ref{fig:modeling}.
The pruning points detection performance depends on both the segmentation inference and the plant modeling algorithm.
This algorithm relies on a set of assessments, for which it uses heuristics reasoning, therefore its performance is affected by how these assessments are implemented.
Furthermore, being strongly dependent on the depth information, its performance heavily depends on the type of the depth sensor, as we tackle small-scale fine-grained objects, i.e. grapevine organs, which can be thin and rough, causing problems in the distance estimation.
However, thanks to the use of redundant reasoning, as discussed above, it is able to compute pruning points even if items, such as \textit{nodes}, are not detected.

\section{Conclusions}
\label{sec:conclusions}
In this work, we presented a new expert annotated dataset for grapevine segmentation, a state of the art neural network training and testing and generation of pruning points that follow agronomic rules.
The dataset is publicly available in open access format at \url{https://zenodo.org/record/5501784}.
We also have shown how the new five-class categorization, the use of more constraining types of relationships and the use of the defined set of assessments allowed us to develop a system that is capable to estimate a set of significant pruning points.
A potential way to improve the segmentation network performance is to use depth information for background removal.
Regarding the grapevine modeling algorithm, its limitations are mainly related to the occlusions that can not be examined from a single point of view.

\bibliographystyle{IEEEtran}

\bibliography{root}

\end{document}